\documentclass{article}
\usepackage{spconf,amsmath,graphicx}

\usepackage{url}
\usepackage{hyperref}
\usepackage[capitalize,noabbrev]{cleveref}
\usepackage{xcolor}

\usepackage[normalem]{ulem}

\def\footurl#1{\footnote{\url{#1}}}

\usepackage[usestackEOL]{stackengine}

\usepackage{diagbox}

\usepackage{multirow}
\usepackage{multicol}
\usepackage{pifont}

\title{Code-Switching without Switching:\\Language Agnostic End-to-End Speech Translation}
\name{Christian Huber$^1$, Enes Yavuz Ugan$^1$, and Alexander Waibel$^{1,2}$}
\address{
  $^1$Interactive Systems Lab, Karlsruhe Institute of Technology, Karlsruhe, Germany\\
  $^2$Carnegie Mellon University, Pittsburgh PA, USA\\
  firstname.lastname@kit.edu, alexander.waibel@cmu.edu}
\begin{document}
%
\maketitle
\begin{abstract}

We propose a) a Language Agnostic end-to-end Speech Translation model (LAST), and b) a data augmentation strategy to increase code-switching (CS) performance.

With increasing globalization, multiple languages are increasingly used interchangeably during fluent speech. Such CS complicates traditional speech recognition and translation, as we must recognize which language was spoken first and then apply a language-dependent recognizer and subsequent translation component to generate the desired target language output. Such a pipeline introduces latency and errors. In this paper, we eliminate the need for that, by treating speech recognition and translation as one unified end-to-end speech translation problem. By training LAST with both input languages, we decode speech into one target language, regardless of the input language. LAST delivers comparable recognition and speech translation accuracy in monolingual usage, while reducing latency and error rate considerably when CS is observed.

\end{abstract}

\section{Introduction}

Due to increasing globalization,
multiple languages are increasingly used interchangeably during fluent speech. 
This is referred to as code-switching (CS).

From a linguistic perspective, CS can be divided into multiple categories \cite{poplack1980sometimes}:
\begin{itemize}
    \item Inter-sentential CS: The switch between languages happens at sentence boundaries. Usually, the speaker is aware of the language shift.
    \item Intra-sentential CS: Here the second language is included in the middle of the sentence. This switch mainly occurs unaware of the speaker. Additionally, the word borrowed from the second language can happen to be adapted to the grammar of the first language as well.
    \item Extra-sentential CS: In this case, a tag element from a second language is included, for example at the end of a sentence. This word is more excluded from the main language.
\end{itemize}

\begin{figure}[t]
    \centering
    \includegraphics[trim={4.95cm 1.4cm 14.4cm 5.8cm},clip,width=1.0\linewidth]{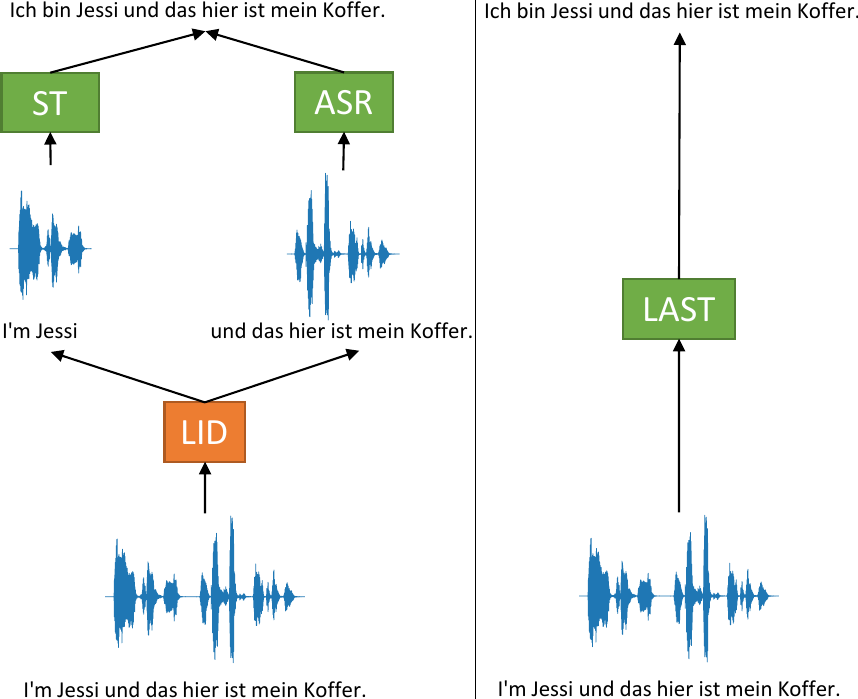}
    \caption{Illustration of the information flow. Left: Baseline: Language identification (LID) followed by either the speech translation (ST) or automatic speech recognition (ASR) model. Right: Our LAST approach. For the green boxes, we use a transformer based encoder-decoder model. The transcripts in the middle and at the bottom are only shown for illustration purposes (the models don't have access to it).}
    \label{fig:figure}
\end{figure}

As of today, there are only a few CS datasets.
Some example corpora available are \cite{amazouz2017addressing} for CS between French and Algerian speech, SEAME from \cite{lyu2010analysis} containing utterances switching between Mandarin and English and \cite{chan2005development} gathered data with CS between English and Cantonese. The Fisher CS dataset \cite{cieri2004fisher} and the Bangor Miami CS dataset \cite{deuchar20145} contain CS automatic speech recognition (ASR) transcripts in English and Spanish and their translations.

Since these datasets are limited in size and available languages, we instead train our model with data not containing CS and focus mostly on inter-sentential CS.

Our contributions are the following:\\
To deal with inter-sentential CS, instead of 
recognizing which language was spoken first, and then apply a language-dependent recognizer and subsequent translation component (or apply a speech recognition or speech translation component as in figure \ref{fig:figure}, left),
we propose to use a \textbf{L}anguage \textbf{A}gnostic end-to-end \textbf{S}peech \textbf{T}ranslation model (LAST) model 
which treats speech recognition and speech translation as one unified end-to-end speech translation problem (see figure \ref{fig:figure}, right). By training with both input languages, we decode speech into one output target language, regardless of whether it represents input speech from the same or a different language.
The unified system delivers comparable recognition and speech translation accuracy in monolingual usage, while reducing latency and error rate considerably when CS is observed.
Furthermore, the pipeline is simplified considerably.

This is shown by evaluating on a testset denoted \emph{tst-inter}. We created this testset, in which the audio contains language switches, for language agnostic speech translation from tst-COMMON. While performing comparable on ASR and speech translation (ST) testsets, LAST increases performance by 7.3 BLEU on tst-inter, compared to a human-annotated LID followed by an ASR or ST model.
Furthermore, we use a data augmentation strategy to increase performance for utterances which have multiple input languages.
With the data augmentation strategy of concatenating audio and corresponding labels of multiple utterances with different source languages into one new utterance, performance of LAST increases by 3.3 BLEU on tst-inter.

The paper is structured as follows: In the following chapter we look at related work, in chapter \ref{sec:exre} we report the used data, model, results and limitations, and in chapter \ref{sec:conc} we conclude our results.

\section{Related Work}


Since there are only a few CS datasets available, there has not been too much research for language pairs without such data.
\cite{seki2018end} propose a model which has the union of graphemes of all languages plus language-specific tags as the target label.
In order to gain performance in the task of CS, they suggest artificially generating training data that contains CS utterances. 
In order to achieve this, they combine full-length utterances of different languages. 
When concatenating the corresponding targets the language-specific token is also added before the target sequence of the respective utterance. 
For our LAST approach, this is not necessary, since we have only one language in the label.
The authors of \cite{yilmaz2018acoustic} used ASR models with a separate TDNN-LSTM \cite{peddinti2017low} as an acoustic model, as well as a separate language model. 
Thus they are able to utilize CS speech-only data for enhancing the acoustic model and used CS text-only data they artificially created, using different approaches, for enhancing their language model.

Most of the work on CS, however, focuses on language pairs where some transcribed CS data is available.
In \cite{shan2019investigating} the authors aim at improving CS performance using a multi-task learning (MTL) approach.
The authors investigate training a model predicting a sequence of labels as well as predicting a language identifier at different levels. 
\cite{shah2020learning} propose to use the Learning without Forgetting \cite{li2017learning} framework to adapt a on monolingual data trained model to CS.
In \cite{li2019towards} the authors propose to train a CTC model \cite{graves2006connectionist} for speech recognition and to linearly adjust the posteriors using a frame-level language identification model. 
The authors of \cite{zhang2022reducing} modify the self-attention of the decoder to reduce multilingual context confusion and to improve the performance of the CS ASR model.

Most similar to our work is the model E2E BIDIRECT SHARED of \cite[figure 3G]{weller2022end}.
However, the difference to our work is, that \cite{weller2022end} uses CS data, where they need transcriptions and translations, as well as annotations which words are from which language, and they focus on intra-sentential CS. Furthermore, they first generate a transcription and therefore have to explicitly detect which language is spoken in each part of the audio. Errors in the transcription step can lead to worse translation performance.

\begin{table}[h]
\begin{tabular}{|l|r|r|} \hline
 Corpus & Utterances & Speech data [h]\\ \hline\hline
 \textbf{A: Training Data: ASR} & 949k & 1825\\
 \hline\hline
Europarl & 64k & 148\\
Librivox & 225k & 512\\
Common Voice & 511k & 685\\
LT & 149k & 480\\
\hline\hline
 \textbf{B: Training Data: ST} & 1196 & 1995\\
 \hline\hline
MuST-C v1 & 230k & 400\\
MuST-C v2 & 251k & 450\\
Europarl-ST & 33k & 77\\
ST TED & 142k & 210\\
CoVoST v2 & 272k & 404\\
TED LIUM & 268k & 454\\
\hline\hline
 \multicolumn{3}{|l|}{\textbf{C: Test Data}} \\
 \hline\hline
tst-COMMON (EN to DE) & 2580 & 4.2\\
tst2013 (DE to DE) & 1369 & 1.9\\
tst2014 (DE to DE) & 1414 & 2.5\\
tst2015 (DE to DE) & 4486 & 3.0\\
tst-inter (EN and DE to DE) & 284 (746) & 0.9\\
\hline
\end{tabular}
\caption{\label{table:st-data} Summary of the datasets used for training. The tst-inter dataset we created contains 746 segments when splitting by the languages.}
\end{table}

\section{Experiments and Results}
\label{sec:exre}

\subsection{Data}

For training and evaluation of our models, we use the German ASR datasets Europarl \cite{iranzo2020europarl}, Librivox \cite{beilharz19}, Common Voice \cite{ardila2019common} and
an internal Lecture Translator (LT) dataset, containing transcribed speech from lectures at KIT,
and the English to German ST datasets MuST-C v1, MuST-C v2 \cite{mustc19}, Europarl-ST \cite{iranzo2020europarl}, ST TED and CoVoST \cite{wang2020covost}, and TED LIUM.
The data split is presented in table \ref{table:st-data}.

Since some of the datasets do not contain casing or punctuation, we trained a transformer encoder \cite{vaswani2017attention} based model to automatically infer this information. The input text is represented as byte-pair-encoding (BPE) \cite{kudo2018sentencepiece} and for each position which is at the start of a word, it is learned if the word should be capitalized and if some punctuation should occur after the word. All the German Wikipedia text is used for training the model.

Furthermore, to evaluate performance of our models when there are multiple languages in the input audio, we derived a test dataset denoted by \emph{tst-inter}\footnote{tst-inter is available for research purposes \href{https://isl-media.anthropomatik.kit.edu/lto/data_share/tst-inter.tar.gz
}{here}.} from tst-COMMON. We looked at the English and German transcripts and divided each utterance into parts where a human might switch the language, e.g. after a comma, a full stop or the word "and". Then the text was read, switching between English and German in each utterance. tst-inter contains almost one hour speech, 746 segments in 284 utterances. 178 utterances contain one switch, 59 two switches, 32 three switches and the rest four or more. Half of the utterances begin with English speech and the other half with German speech. We also annotated the language id (LID), i.e., which language is spoken in each part of the audio. 



\subsection{Models}

We use the framework
\href{https://github.com/quanpn90/NMTGMinor}{NMTGMinor} 
which is based on PyTorch and uses the Fairseq pretrained models
for training.
Similar to recent works, e.g., \cite{li2020multilingual}, we start with a transformer model \cite{vaswani2017attention, pham2019very}, where the
encoder is initialized with the pretrained Wav2Vec 2.0 model \cite{baevski2020wav2vec} and the
decoder is initialized with the decoder part of the pretrained mBART 50 model \cite{liu2020multilingual}.
In particular, since we work with multiple input languages, we use the  \href{https://huggingface.co/facebook/wav2vec2-large-xlsr-53}{facebook/wav2vec2-large-xlsr-53} checkpoint for the initialization of Wav2Vec 2.0, and the \href{https://huggingface.co/facebook/mbart-large-50}{facebook/mbart-large-50} checkpoint for the initialization of mBART 50.

We finetuned with our training data for 40k updates with 18k target tokens per update, adam optimizer with a maximum learning rate of 5e-4 and 4k warmup steps. 
We used only utterances with a maximum length 20 seconds and the embedding layer was frozen. After convergence, we averaged the five best epochs according to perplexity on the validation set containing 6k utterances removed from the training data.

With this procedure, we trained three models: For comparison, one with only the ASR data and one with only the ST data, and one with all the data, denoted \textbf{L}anguage \textbf{A}gnostic end-to-end \textbf{S}peech \textbf{T}ranslation (LAST) model.

Furthermore, we use curriculum learning \cite{bengio2009curriculum} and finetuned the LAST model employing additionally the data augmentation (DA) strategy of concatenating audio and labels of multiple utterances with different source languages into one new utterance. We applied the same procedure stated above for 4k updates and trained models with different amount of utterances with multiple input languages.
The percentage reported corresponds to how many of the utterances used, contain at least two input languages in the audio. For these utterances, 80\% are selected to switch the language once, 20\% are selected to switch the language twice. The percentage of 75\% is the maximum we could achieve by allowing a maximum of 20 seconds of audio.

Note, that the LAST model we trained gets as input English and German audio, and we decode both of these input languages into German text (see figure \ref{fig:figure}, right). 
Therefore, by treating speech recognition and speech translation as one unified end-to-end speech translation problem, we eliminate the need for a language switch in the decoder.

\subsection{Results}

In table \ref{table:results} the results of the ASR and ST testsets can be seen. We report the WER on tst2013-tst2015 and the sacreBLEU score on tst-COMMON.
We can see that the ASR and ST performance of the LAST model as well as the LAST models with data augmentation is comparable to the baselines (on two testsets slightly better, on two testsets slightly worse), even though the LAST models are able to handle both tasks.

\begin{table}[t]
  \centering
  \resizebox{1.0\columnwidth}{!}{%
    \begin{tabular}{|r|c|c|c|c|} \hline
    \multicolumn{1}{|r|}{\backslashbox{Model}{Dataset}} & tst2013 & tst2014 & tst2015 & \Centerstack{tst-\\COMMON}\\
    \hline\hline
    ASR & 14.3 & \textbf{11.0} & \textbf{10.0} & --\\
    ST & -- & -- & -- & 30.9\\
    \hline
    LAST\qquad\qquad& 13.9 & 11.4 & 10.7 & \textbf{31.1}\\
    \qquad+DA \ \ 5\% & \textbf{13.4} & 11.1 & 10.3 & 30.9\\
    \qquad+DA 10\% & 13.6 & 11.4 & 10.5 & 30.9\\
    \qquad+DA 15\% & 13.5 & 11.3 & 10.3 & 30.8\\
    \qquad+DA 20\% & 13.7 & 11.3 & 10.3 & 30.7\\
    \qquad+DA 30\% & 13.6 & 11.4 & 10.5 & 30.9\\
    \qquad+DA 40\% & \textbf{13.4} & 11.3 & 10.4 & 30.8\\
    \qquad+DA 75\% & 13.6 & 11.4 & 10.8 & 30.9\\
    \hline    LAST half data & 14.6 & 11.8 & 11.0 & 30.7\\
    \hline
    \end{tabular}
  }
\caption{\label{table:results} Summary of the monolingual ASR and ST results. WER ($\downarrow$) on tst2013, tst2014 and tst2015 and sacreBLEU score ($\uparrow$) on tst-Common as metrics. First two rows: Baselines, last row: For comparison, other rows: Our method. }
\end{table}

Since LAST is trained with double the data compared to the ASR or ST models, we also compare to \emph{LAST half data} which is the LAST model trained with half the training data. 
We obtain that this model is a bit worse than the other models, which was expectable due to the reduction in training data.

\begin{table}[t]
  \centering
    \begin{tabular}{|r|c|c|c|c|} \hline
    \multicolumn{1}{|r|}{\backslashbox{Model}{Metric}} & \Centerstack{sacreBLEU}& \Centerstack{+ no punct}\\
    \hline\hline
    given LID + ASR or ST & 38.6 & 43.4\\
    \hline
    given LID + LAST & 39.1 & 43.6\\
    \hline
    LAST half data & 45.3 & 45.5\\
    \hline
    LAST\qquad\qquad& 45.9 & 46.3\\
    \qquad+DA \ \ 5\% & \textbf{49.2} & 49.3\\
    \qquad+DA 10\% & 49.0 & 49.4\\
    \qquad+DA 15\% & \textbf{49.2} & 49.6\\
    \qquad+DA 20\% & 48.7 & 49.4\\
    \qquad+DA 30\% & 49.1 & \textbf{49.7}\\
    \qquad+DA 40\% & \textbf{49.2} & \textbf{49.7}\\
    \qquad+DA 75\% & 48.5 & \textbf{49.7}\\
    \hline
    \end{tabular}
\caption{\label{table:results2} Summary of the results on the bilingual tst-inter testset. sacreBLEU score ($\uparrow$) and additionally sacreBLEU score with removed punctuation on tst-inter. First row: Baseline, second and third row: For comparison, other rows: Our method.
}
\end{table}

In table \ref{table:results2} the results of the tst-inter testset can be seen. The sacreBLEU scores are rather high, since the testset contains parts where the model has to do ASR, in contrast to tst-COMMON, where the task is only ST.
We compare to the baseline \emph{given LID + ASR or ST} (see figure \ref{fig:figure}, left), where we use the given (human-annotated) LID information and split the audio accordingly. Then, we run the ASR or ST model on the segments, depending on the LID information. Finally, we concatenate the outputs.
Note, that each LID model one would run in practice, is expected to make errors and therefore lead to worse performance.




We obtain, that the performance of the LAST model is 7.3 BLEU better than the baseline, even though it does not use the (human-annotated) LID information. When looking at the output, we saw that the baseline produces errors with the punctuation at the positions where the outputs are concatenated (see for example table \ref{table:example}). It is not easily possible to correct them with a post processing step since at the positions of the switches it is possible to have different punctuation. Therefore, we evaluated the sacreBLEU score with removed punctuation in hypothesis and reference, and see the same trends.
The LAST model might perform better than the baseline because it has access to more context (see figure \ref{fig:figure} for example). The ST or ASR models of the baseline can only be fed with parts of the input containing one specific language. Using the ST or ASR models with the whole sequence would result in drastically worse performance since these models have only seen one language during training. In contrast, the LAST model can use the full input audio sequence.

For comparison, we also report \emph{given LID + LAST}, which is similar to the baseline, but instead of running the ASR or ST model, the LAST model is used. We see, that this approach slightly increases the performance on tst-inter. However, there is still a huge gap to the performance of the LAST models.
Furthermore, we see that \emph{LAST half data} performs only slightly worse than the LAST model on tst-inter. Therefore, the improvement of our method is not due to more data but more available context as stated above.

When looking at the results for the models with the additional data augmentation,
we see the following: Which percentage to use has only limited effect on the results (as long as the percentage is larger than zero), but the best model improves 3.3 BLEU over the LAST model without this data augmentation and 10.6 BLEU over the baseline. Therefore, this data augmentation strategy heavily boosts performance on this testset.


Note that the LAST models (with or without data augmentation) are able to be applied in an online low latency setup where language switches occur. Compared to the baseline, it is not necessary to run some LID system and then the speech recognition or speech translation model. This reduces latency since there is no pipeline which has to do discrete decisions. Additionally, from an implementation perspective, LAST is easier to deploy/maintain, because it consists of fewer components.

\begin{table}
\centering
  \resizebox{1.0\columnwidth}{!}{%
    \begin{tabular}{|r|l|}
    \hline
    Reference & \Centerstack{Aber bevor ich Ihnen zeige,\\ was ich darin habe,\\ werde ich ein sehr öffentliches Geständnis machen,\\ und das ist:\\ ich bin besessen von Outfits.}\\
    \hline\hline
    given LID + ASR or ST & \Centerstack{Aber bevor ich Ihnen zeige,\\ was ich darin habe.\\ ich werde ein sehr öffentliches Geständnis ablegen.\\ Und das ist?\\ ich bin vom Outfit besessen.}\\
    \hline
    given LID + LAST & \Centerstack{Aber bevor ich Ihnen zeige,\\ was ich darin habe?\\ ich werde ein sehr öffentliches Geständnis ablegen.\\ und das ist\\ ich bin vom Outfit besessen.}\\
    \hline
    LAST\,\, half data & \Centerstack{Aber bevor ich Ihnen zeige,\\ was ich darin habe,\\ werde ich ein sehr öffentliches Geständnis ablegen,\\ und das ist\\ "I am Outfit besessen".}\\
    \hline
    LAST\qquad\qquad\  & \Centerstack{Aber bevor ich Ihnen zeige,\\ was ich darin habe,\\ werde ich ein sehr öffentliches Geständnis ablegen,\\ und das ist\\ "I am Outfit" besessen.}\\ \cline{2-2}
    \qquad+DA \ \ 5\% & \Centerstack{Aber bevor ich Ihnen zeige,\\ was ich darin habe,\\ werde ich ein sehr öffentliches Geständnis ablegen,\\ und das ist:\\ ich bin "outfit" besessen.}\\ \cline{2-2}
    \qquad+DA 10\%/15\% & \Centerstack{Aber bevor ich Ihnen zeige,\\ was ich darin habe,\\ werde ich ein sehr öffentliches Geständnis ablegen,\\ und das ist,\\ ich bin "outfit" besessen.}\\ \cline{2-2}
    \qquad+DA 20\% & \Centerstack{Aber bevor ich Ihnen zeige,\\ was ich darin habe,\\ werde ich ein sehr öffentliches Geständnis ablegen,\\ und das ist:\\ ich bin "outfit-besessen".}\\ \cline{2-2}
    \qquad+DA 30\% & \Centerstack{Aber bevor ich Ihnen zeige,\\ was ich darin habe,\\ werde ich ein sehr öffentliches Geständnis ablegen,\\ und das ist,\\ ich bin "outfit-besessen".}\\ \cline{2-2}
    \qquad+DA 40\% & \Centerstack{Aber bevor ich Ihnen zeige,\\ was ich darin habe,\\ werde ich ein sehr öffentliches Geständnis ablegen,\\ und das ist,\\ ich bin vom Outfit besessen.}\\ \cline{2-2}
    \qquad+DA 75\% & \Centerstack{Aber bevor ich Ihnen zeige,\\ was ich darin habe,\\ werde ich ein sehr öffentliches Geständnis ablegen,\\ und das ist\\ "I Am Outfit Obsessed".}\\
    \hline\hline
    Transcript & \Centerstack{Aber bevor ich Ihnen zeige,\\ was ich darin habe,\\ I'm going to make a very public confession,\\ und das ist:\\ I'm outfit-obsessed.}\\
    \hline
    \end{tabular}
  }
\caption{\label{table:example} Example output of an utterance in tst-inter for all models. 
For comparison we also report the transcript of the audio.
}
\end{table}

\subsection{Limitations}


We reviewed test examples of intra-sentential CS in German speech (containing English words) qualitatively, but could not see any improvement of the LAST model compared to the ASR model. An explanation could be that neither the LAST model nor the ASR model have seen intra-sentential CS during training. Further research in this area is required.

\section{Conclusion}
\label{sec:conc}

In this work, we proposed a Language Agnostic Speech Translation model 
which treats speech recognition and speech translation as one unified end-to-end speech translation problem. By training with both input languages, we decode speech into one output target language, regardless of whether it represents input speech from the same or a different language. The unified system delivers comparable recognition and speech translation accuracy in monolingual usage, while reducing latency and error rate considerably when CS is observed.



\newpage
\section{Acknowledgements}

We want to thank Ngoc Quan Pham for providing his framework NMTGMinor to train our systems.
The projects on which this paper is based were funded by the Federal Ministry of Education and Research (BMBF) of Germany under the numbers 01IS18040A (OML) and 01EF1803B (RELATER).


\bibliographystyle{IEEEbib}
\bibliography{refs}

\begin{thebibliography}{10}

\bibitem{poplack1980sometimes}
Shana Poplack,
\newblock ``Sometimes i’ll start a sentence in spanish y termino en espanol:
  toward a typology of code-switching1,''
\newblock 1980.

\bibitem{amazouz2017addressing}
Djegdjiga Amazouz, Martine Adda-Decker, and Lori Lamel,
\newblock ``Addressing code-switching in french/algerian arabic speech,''
\newblock in {\em Interspeech 2017}, 2017, pp. 62--66.

\bibitem{lyu2010analysis}
Dau-Cheng Lyu, Tien-Ping Tan, Eng-Siong Chng, and Haizhou Li,
\newblock ``An analysis of a mandarin-english code-switching speech corpus:
  Seame,''
\newblock {\em Age}, vol. 21, pp. 25--8, 2010.

\bibitem{chan2005development}
Joyce~YC Chan, PC~Ching, and Tan Lee,
\newblock ``Development of a cantonese-english code-mixing speech corpus,''
\newblock in {\em Ninth European Conference on Speech Communication and
  Technology}, 2005.

\bibitem{cieri2004fisher}
Christopher Cieri, David Miller, and Kevin Walker,
\newblock ``The fisher corpus: A resource for the next generations of
  speech-to-text.,''
\newblock in {\em LREC}, 2004, vol.~4, pp. 69--71.

\bibitem{deuchar20145}
Margaret Deuchar, Peredur Davies, Jon~Russell Herring, M~Carmen~Parafita Couto,
  and Diana Carter,
\newblock ``5. building bilingual corpora,''
\newblock in {\em Advances in the Study of Bilingualism}, pp. 93--110.
  Multilingual Matters, 2014.

\bibitem{seki2018end}
Hiroshi Seki, Shinji Watanabe, Takaaki Hori, Jonathan Le~Roux, and John~R
  Hershey,
\newblock ``An end-to-end language-tracking speech recognizer for
  mixed-language speech,''
\newblock in {\em 2018 IEEE International Conference on Acoustics, Speech and
  Signal Processing (ICASSP)}. IEEE, 2018, pp. 4919--4923.

\bibitem{yilmaz2018acoustic}
Emre Y{\i}lmaz, Henk van~den Heuvel, and David~A van Leeuwen,
\newblock ``Acoustic and textual data augmentation for improved asr of
  code-switching speech,''
\newblock {\em arXiv preprint arXiv:1807.10945}, 2018.

\bibitem{peddinti2017low}
Vijayaditya Peddinti, Yiming Wang, Daniel Povey, and Sanjeev Khudanpur,
\newblock ``Low latency acoustic modeling using temporal convolution and
  lstms,''
\newblock {\em IEEE Signal Processing Letters}, vol. 25, no. 3, pp. 373--377,
  2017.

\bibitem{shan2019investigating}
Changhao Shan, Chao Weng, Guangsen Wang, Dan Su, Min Luo, Dong Yu, and Lei Xie,
\newblock ``Investigating end-to-end speech recognition for mandarin-english
  code-switching,''
\newblock in {\em ICASSP 2019-2019 IEEE International Conference on Acoustics,
  Speech and Signal Processing (ICASSP)}. IEEE, 2019, pp. 6056--6060.

\bibitem{shah2020learning}
Sanket Shah, Basil Abraham, Sunayana Sitaram, Vikas Joshi, et~al.,
\newblock ``Learning to recognize code-switched speech without forgetting
  monolingual speech recognition,''
\newblock {\em arXiv preprint arXiv:2006.00782}, 2020.

\bibitem{li2017learning}
Zhizhong Li and Derek Hoiem,
\newblock ``Learning without forgetting,''
\newblock {\em IEEE transactions on pattern analysis and machine intelligence},
  vol. 40, no. 12, pp. 2935--2947, 2017.

\bibitem{li2019towards}
Ke~Li, Jinyu Li, Guoli Ye, Rui Zhao, and Yifan Gong,
\newblock ``Towards code-switching asr for end-to-end ctc models,''
\newblock in {\em ICASSP 2019-2019 IEEE International Conference on Acoustics,
  Speech and Signal Processing (ICASSP)}. IEEE, 2019, pp. 6076--6080.

\bibitem{graves2006connectionist}
Alex Graves, Santiago Fern{\'a}ndez, Faustino Gomez, and J{\"u}rgen
  Schmidhuber,
\newblock ``Connectionist temporal classification: labelling unsegmented
  sequence data with recurrent neural networks,''
\newblock in {\em Proceedings of the 23rd international conference on Machine
  learning}, 2006, pp. 369--376.

\bibitem{zhang2022reducing}
Shuai Zhang, Jiangyan Yi, Zhengkun Tian, Jianhua Tao, Yu~Ting Yeung, and Liqun
  Deng,
\newblock ``Reducing language context confusion for end-to-end code-switching
  automatic speech recognition,''
\newblock {\em arXiv preprint arXiv:2201.12155}, 2022.

\bibitem{weller2022end}
Orion Weller, Matthias Sperber, Telmo Pires, Hendra Setiawan, Christian Gollan,
  Dominic Telaar, and Matthias Paulik,
\newblock ``End-to-end speech translation for code switched speech,''
\newblock {\em arXiv preprint arXiv:2204.05076}, 2022.

\bibitem{iranzo2020europarl}
Javier Iranzo-S{\'a}nchez, Joan~Albert Silvestre-Cerda, Javier Jorge, Nahuel
  Rosell{\'o}, Adria Gim{\'e}nez, Albert Sanchis, Jorge Civera, and Alfons
  Juan,
\newblock ``Europarl-st: A multilingual corpus for speech translation of
  parliamentary debates,''
\newblock in {\em ICASSP 2020-2020 IEEE International Conference on Acoustics,
  Speech and Signal Processing (ICASSP)}. IEEE, 2020, pp. 8229--8233.

\bibitem{beilharz19}
Benjamin Beilharz, Xin Sun, Sariya Karimova, and Stefan Riezler,
\newblock ``Librivoxdeen: A corpus for german-to-english speech translation and
  speech recognition,''
\newblock {\em Proceedings of the Language Resources and Evaluation
  Conference}, 2020.

\bibitem{ardila2019common}
Rosana Ardila, Megan Branson, Kelly Davis, Michael Henretty, Michael Kohler,
  Josh Meyer, Reuben Morais, Lindsay Saunders, Francis~M Tyers, and Gregor
  Weber,
\newblock ``Common voice: A massively-multilingual speech corpus,''
\newblock {\em arXiv preprint arXiv:1912.06670}, 2019.

\bibitem{mustc19}
Mattia~A. Di~Gangi, Roldano Cattoni, Luisa Bentivogli, Matteo Negri, and Marco
  Turchi,
\newblock ``{M}u{ST}-{C}: a {M}ultilingual {S}peech {T}ranslation {C}orpus,''
\newblock in {\em Proceedings of the Conference of the North {A}merican Chapter
  of the Association for Computational Linguistics (NAACL)}, June 2019.

\bibitem{wang2020covost}
Changhan Wang, Anne Wu, and Juan Pino,
\newblock ``Covost 2 and massively multilingual speech-to-text translation,''
\newblock {\em arXiv preprint arXiv:2007.10310}, 2020.

\bibitem{vaswani2017attention}
Ashish Vaswani, Noam Shazeer, Niki Parmar, Jakob Uszkoreit, Llion Jones,
  Aidan~N Gomez, {\L}ukasz Kaiser, and Illia Polosukhin,
\newblock ``Attention is all you need,''
\newblock {\em Advances in neural information processing systems}, vol. 30,
  2017.

\bibitem{kudo2018sentencepiece}
Taku Kudo and John Richardson,
\newblock ``Sentencepiece: A simple and language independent subword tokenizer
  and detokenizer for neural text processing,''
\newblock {\em arXiv preprint arXiv:1808.06226}, 2018.

\bibitem{li2020multilingual}
Xian Li, Changhan Wang, Yun Tang, Chau Tran, Yuqing Tang, Juan Pino, Alexei
  Baevski, Alexis Conneau, and Michael Auli,
\newblock ``Multilingual speech translation with efficient finetuning of
  pretrained models,''
\newblock {\em arXiv preprint arXiv:2010.12829}, 2020.

\bibitem{pham2019very}
Ngoc-Quan Pham, Thai-Son Nguyen, Jan Niehues, Markus M{\"u}ller, Sebastian
  St{\"u}ker, and Alexander Waibel,
\newblock ``Very deep self-attention networks for end-to-end speech
  recognition,''
\newblock {\em arXiv preprint arXiv:1904.13377}, 2019.

\bibitem{baevski2020wav2vec}
Alexei Baevski, Yuhao Zhou, Abdelrahman Mohamed, and Michael Auli,
\newblock ``wav2vec 2.0: A framework for self-supervised learning of speech
  representations,''
\newblock {\em Advances in Neural Information Processing Systems}, vol. 33, pp.
  12449--12460, 2020.

\bibitem{liu2020multilingual}
Yinhan Liu, Jiatao Gu, Naman Goyal, Xian Li, Sergey Edunov, Marjan
  Ghazvininejad, Mike Lewis, and Luke Zettlemoyer,
\newblock ``Multilingual denoising pre-training for neural machine
  translation,''
\newblock {\em Transactions of the Association for Computational Linguistics},
  vol. 8, pp. 726--742, 2020.

\bibitem{bengio2009curriculum}
Yoshua Bengio, J{\'e}r{\^o}me Louradour, Ronan Collobert, and Jason Weston,
\newblock ``Curriculum learning,''
\newblock in {\em Proceedings of the 26th annual international conference on
  machine learning}, 2009, pp. 41--48.

\end{thebibliography}

\end{document}